\pdfoutput=1
\documentclass[11pt]{article}

\usepackage[review]{ACL2023}

\usepackage{times}
\usepackage{latexsym}

\usepackage[T1]{fontenc}
\usepackage[utf8]{inputenc}

\usepackage{microtype}

\usepackage{inconsolata}
\usepackage{CJKutf8}
\usepackage{caption}
\usepackage{graphicx}
\usepackage{float} 
\usepackage{subcaption}
\usepackage{multirow}
\usepackage{amssymb}
\usepackage{amsmath, amssymb, mathtools}
\usepackage{booktabs}

\title{IEKM: A Model Incorporating External Keyword Matrices}

\author{Cheng Luo\textsuperscript{1}, Qin Li\textsuperscript{1}, Zhao Yan\textsuperscript{1}, Mengliang Rao\textsuperscript{1} \and Yunbo Cao\textsuperscript{1} \\
\textsuperscript{1}Tencent \\
\{luochengluo, jemmali, zhaoyan, sekarao, yunbocao\}@tencent.com
}
\begin{document}
\maketitle
\begin{abstract}
A customer service platform system with a core text semantic similarity (STS) task faces two urgent challenges: Firstly, one platform system needs to adapt to different domains of customers, i.e., different domains adaptation (DDA). 
Secondly, it is difficult for the model of the platform system to distinguish sentence pairs that are literally close but semantically different, i.e., hard negative samples.
In this paper, we propose an incorporation external keywords matrices model (IEKM) to address these challenges.
The model uses external tools or dictionaries to construct external matrices and fuses them to the self-attention layers of the Transformer structure through gating units, thus enabling flexible corrections to the model results.
We evaluate the method on multiple datasets and the results show that our method has improved performance on all datasets.
To demonstrate that our method can effectively solve all the above challenges,
we conduct a flexible correction experiment, which results in an increase in the F1 value from 56.61 to 73.53.
Our code will be publicly available.
\end{abstract}

\section{Introduction}
% 讲故事
% 背景
Our work focuses on a real industrial scenario, i.e. a retrieval-based customer service platform system.
When faced a query from customer, the system first uses a coarse-grained algorithm (BM25 \citep{DBLP:journals/ftir/RobertsonZ09}) to recall the top-n relevant candidates for the query and then uses a fine-ranking algorithm to sort these candidates to find the most similar sentence to the query to return the most relevant answer \citep{DBLP:conf/emnlp/KarpukhinOMLWEC20}.
% The core task of a retrieval-based customer service system platform is the semantic text similarity (STS) task.
Fine-ranking is the text semantic similarity task, which is the core of the retrieval-based customer service system platform.
% This is also the core task of intelligent customer service, which is to find the most similar sentences to the input corpus from the conversation corpus to get the corresponding response.
% A common industry practice is to first recall candidates by a coarse-grained algorithm and then use a fine-grained matching algorithm to fine-rank the candidates \citep{DBLP:conf/emnlp/KarpukhinOMLWEC20}.
% The coarse-grained algorithm usually uses BM25 \citep{DBLP:journals/ftir/RobertsonZ09}, while the fine-grained matching algorithm usually uses a pre-trained bidirectional encoder such as BERT \citep{DBLP:conf/naacl/DevlinCLT19}.
And, it is a common phenomenon that the candidates recalled by coarse-grained algorithms often contain a large number of hard negative samples, i.e., the sentences are literally close to the query but semantically different.
For example, the query \begin{CJK*}{UTF8}{gbsn}"\begin{small}我的余额宝里有多少钱？\end{small} (How much money do I have in my Yu'E Bao?)"\end{CJK*} and the candidate \begin{CJK*}{UTF8}{gbsn}"\begin{small}我的余额+里有多少钱？\end{small} (How much money do I have in my Yu'E+?)"\end{CJK*} should be semantically far away, but the model usually considers them semantically close (Yu'E Bao and Yu'E+ are two different products, so the customer requirement is not the same.).
Another common phenomenon in the industry is that one customer service platform system needs to meet the needs of customers in different domains, i.e., different domains adaptation (DDA).
For example, the sentence pair \begin{CJK*}{UTF8}{gbsn}[\begin{small}收益哪里看\end{small}, \begin{small}总收益哪里看\end{small}]\end{CJK*} ([Where to look for earnings, Where to see total earnings]) are considered to mean one thing by the customer in the generic domain, but by the customer in the financial domain to mean different things (in the financial world, total earnings and earnings usually represent different metrics.)

\begin{figure}[!h]
  \centering
  \includegraphics[width=0.45\textwidth]{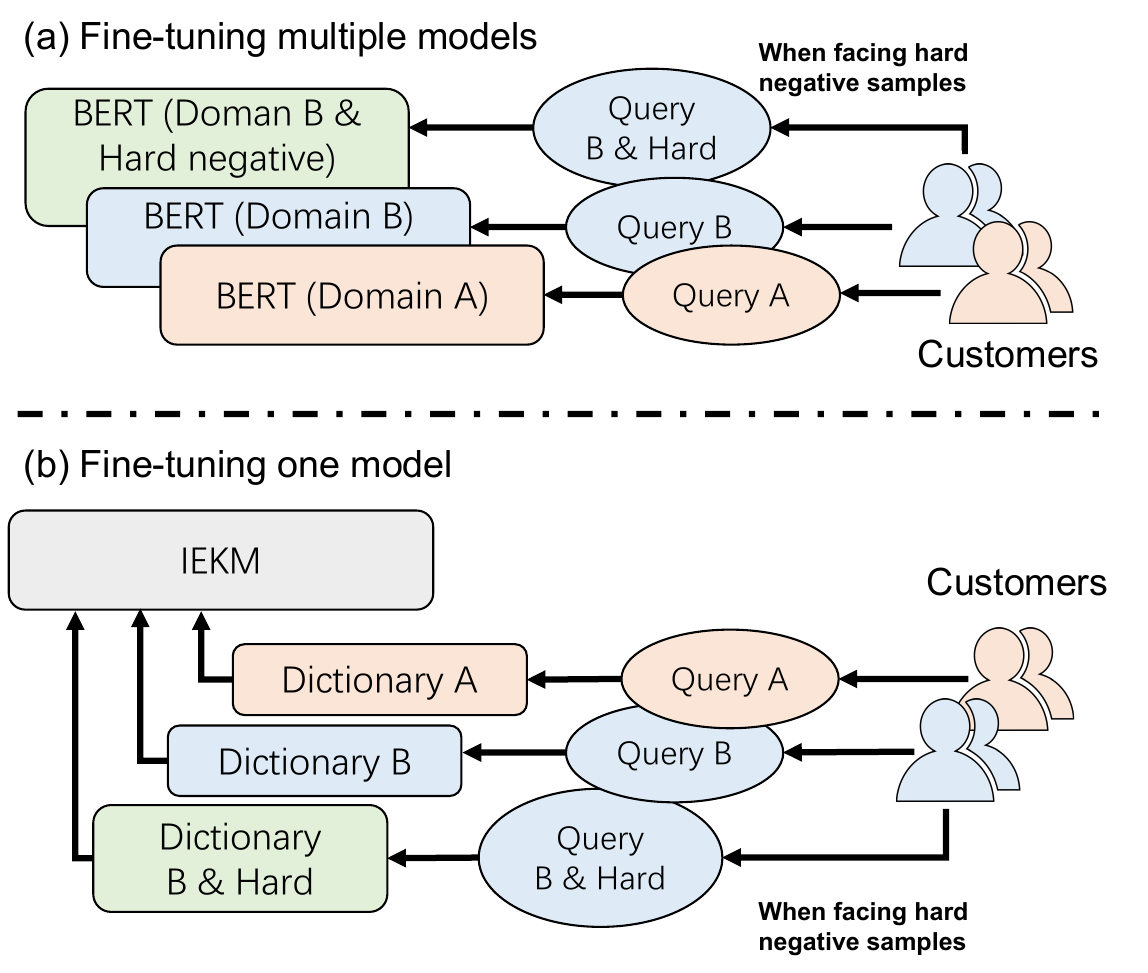}
  \caption{(a) is a common practice in industry to deploy different models for different domain customers. (b) is IEKM's approach to solving DDA challenge. Here we consider the hard negative samples as different domain data as well and different colors represent different domains.}
  \label{fig:model_comper}
\end{figure}

These are two common challenges in STS tasks but are less studied.
In industry, as shown in Figure \ref{fig:model_comper} (a), to address the first challenge above, it is common practice to annotate hard negative samples to retrain the model.
To address the second challenge, different models usually need to be deployed for different domain customers.
These methods are useful but expensive: To obtain enough hard negative data and customer domain data, experts familiar with these scenarios need to be hired to label the data; 
At the same time, deploying different models for different users can take up a lot of hardware resources.
As the number of customers with different needs increases, the amount of data to be labeled and the number of models to be deployed increases, so there is a huge cost overhead in the long run with this method.

%通过对不同领域的数据研究，我们发现领域内的一些关键词对句子语义影响特别大,并更容易获取.
By studying data from different domains, we find that keywords within the domain play a key role in sentence semantics, which is consistent with the observation of \cite{DBLP:conf/acl/PeineltNL20} and \cite{miao2020keyword} et al.
And keywords are easier to obtain than labeled data pairs in real-world.
In order to incorporate this useful external information into the model to address the above challenges, in this paper, we propose IEKM, a method by incorporating external similarity word matrices (Figure \ref{fig:model_comper} (b)).
Specifically, we use the tools for calculating similar word scores to construct positive and negative correlation matrices, respectively. 
Then they are incorporated into the self-attention layers of IEKM (based on Transformer encoder \citep{DBLP:conf/nips/VaswaniSPUJGKP17}) through the learnable gate control units. 
In this way, the model adjusts the gate control units' scores according to the actual inputs and thus decides how much information to obtain from the external matrices to correct the self-attention scores.
In the training model phase, to obtain the similar word scores, we use the word similarity tools Cilin\footnote{\url{https://github.com/ashengtx/CilinSimilarity}} and Hownet\footnote{\url{https://openhownet.thunlp.org/download}} to construct the external matrices.
%%需要重新调整贡献
After the model is deployed, for different domain customers we build exclusive keyword dictionaries for them. 
During inference, we use customer requests to query this dictionary and the word similarity tools to construct the external matrices.
When facing a new domain customer or a new hard negative sample, we only need to `INSERT' the relevant keyword information into the relevant database to meet the demand, no need to retrain the model and redeploy it.

The main contributions of this paper are as follows:

% 具体贡献
\begin{itemize}
\item  We propose a BERT-based external matrices fusion model called IEKM.

\item  We demonstrate the effectiveness of the method, i.e., the results of the model are corrected by external matrices, thus improving the ability of the model to solve hard negative samples and DDA.

% \item  We qualitatively analyze the reasons why IEKM works.
\end{itemize}

\section{Related Work}

Measuring the semantic similarity of two sentences is the core of STS tasks.
% The candidate sentences will be re-ranked based on these semantic text similarities to obtain the most similar sentence.
Deep learning models have a wide range of applications in measuring the semantic text similarity of two sentences.
Earlier, researchers proposed the DSSM model to obtain the embedding of sentences for similarity matching \citep{DBLP:conf/cikm/HuangHGDAH13}.
Meanwhile, the ESIM model with the application of Bi-LSTM and attention mechanism achieves very excellent results in text matching \citep{DBLP:conf/acl/ChenZLWJI17}.
Recently, pre-trained language models and their variants \citep{radford2019language, DBLP:conf/nips/YangDYCSL19, DBLP:journals/corr/abs-1907-11692, DBLP:conf/naacl/DevlinCLT19} have been widely used in the task of measuring semantic text similarity of texts with great success.
Among them, BERT-flow \citep{DBLP:conf/emnlp/LiZHWYL20} and BERT-whitening \citep{DBLP:journals/corr/abs-2103-15316} improve the performance of computing semantic text similarity by post-processing the sentence embedding after model encoding.
SimCSE \citep{DBLP:conf/emnlp/GaoYC21}, on the other hand, uses the Dropout \citep{DBLP:journals/jmlr/SrivastavaHKSS14} mechanism to construct negative samples for contrastive learning to improve the quality of sentence embeddings extracted by the model.

The use of external knowledge can also bring some improvement to the results of text semantic similarity matching than using only requirement and candidate.
External knowledge can be expressed as a knowledge graph \citep{DBLP:conf/emnlp/WangZFC14}, unstructured text \citep{DBLP:conf/aaai/GhazvininejadBC18}, or a descriptive corpus.
ENRIE \citep{DBLP:conf/acl/ZhangHLJSL19} is an improvement of BERT, which uses a large-scale text corpus and knowledge graph.
It incorporates knowledge from the knowledge graph during pre-training to improve the model's ability of coding sentences.
K-BERT \citep{DBLP:conf/aaai/LiuZ0WJD020}, on the other hand, is improved during fine-tuning.
It injects the triples of the knowledge graph into the input sentences to make the information of the input sentences complete, which in turn makes the text matching performance improved.
There is also a work that makes use of word similarity information, which is incorporated into the model as prior knowledge to improve the model performance, called BERT-Sim \citep{DBLP:conf/www/XiaWTC21}.
Another line of work focuses on incorporating keywords into the BERT model to improve re-ranking accuracy \citep{DBLP:journals/corr/abs-2003-11516}.
This work adds a keyword attention layer to the final layer of BERT to enhance the model's ability to capture keywords.
% However, they still cannot solve the challenge presented above.

\section{Proposed Method}

\begin{figure}[!ht]
  \centering
  \includegraphics[width=0.42\textwidth]{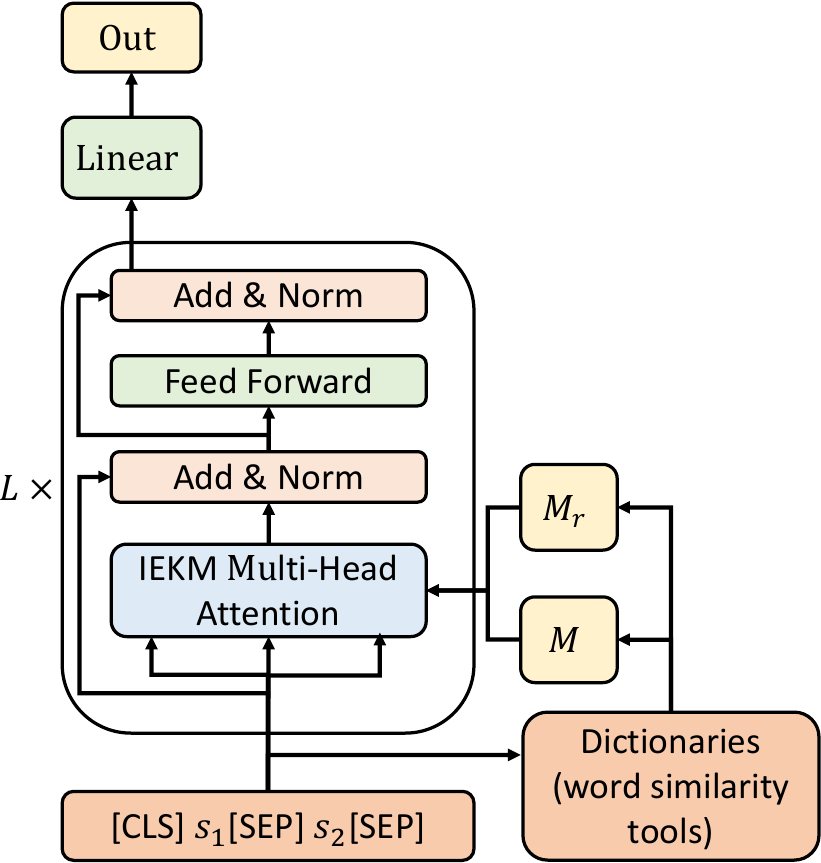}
  \caption{Architecture of IEKM with matrices input, where $s_{1}$, $s_{2}$ are the input sentence pairs. We construct the matrices $M$ and $M_{r}$ using sentence pairs $s_{1}$,$s_{2}$ and input these matrices to the IEKM self-attention layers.}
  \label{fig:architecture}
\end{figure}

\begin{figure*}[htb]
  \centering
  \begin{subfigure}{0.49\textwidth}
      \includegraphics[width=0.8\textwidth]{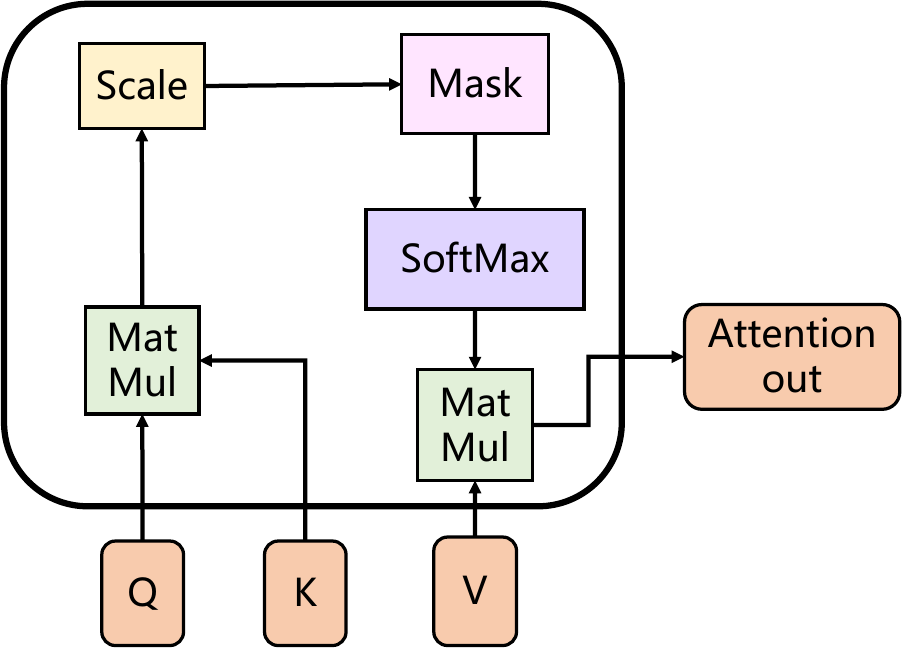}
      \caption{Self-attention.}
    %   \label{fig:ex_matrix}
  \end{subfigure}
    \centering
  \begin{subfigure}{0.5\textwidth}
      \includegraphics[width=1\textwidth]{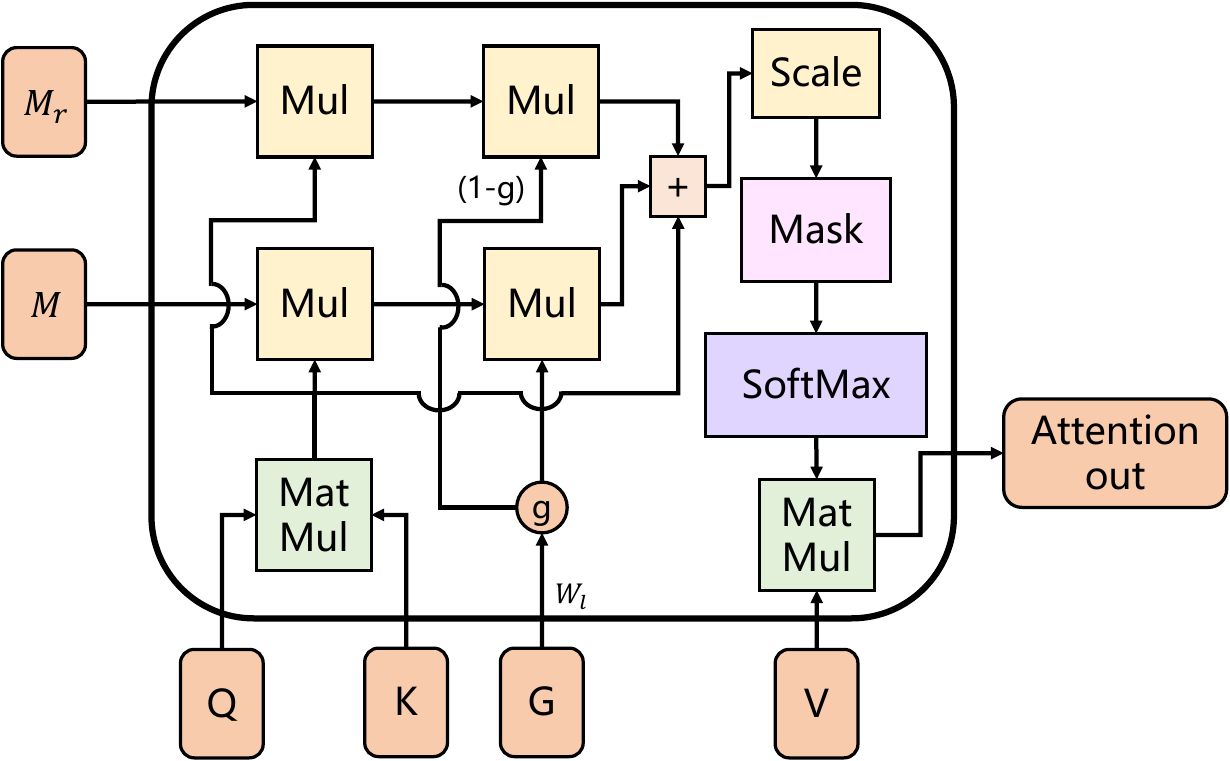}
      \caption{IEKM Self-attention.}
    %   \label{fig:ex_matrix}
  \end{subfigure}
  \caption{Traditional self-attention vs IEKM self-attention in a head.
  % The block with chamfer indicates the data.
  $W$ denotes the learnable parameters in the model.
  $g$ is the gate control unit.}
  \label{fig:Our self-attention}
\end{figure*}

% \begin{figure*}[!ht]
%   \centering
%   \includegraphics[width=0.7\textwidth]{ex_matrix.png}
%   \caption{An example of a similarity matrix $M$. The darker the color of the cell, the larger the score, and the more similar the corresponding tokens are to each other.}
%   \label{fig:ex_matrix}
% \end{figure*}

In our experiments, the training data consists of $\{(S_{i}, W_{i}, Y_{i})\}^N_{i=1}$, where $S=[s_{1}, s_{2}]$ is a sentence pair, $W_{i}=\{c_{11}, c_{12},...,c_{ab}\}^{a=l_{1}, b=l_{2}}_{a, b=1}$ is the set of word similarity scores between sentence pairs $S_{i}$ while $l_{1}$ and $l_{2}$ are the lengths of $s_{1}, s_{2}$ respectively. 
And $Y_{i}$ is the corresponding corresponding sentiment label.
If the training data belongs to the classification task, then $Y \in \{0, 1\}$, but if the training data belongs to the regression task, then $Y \in [0, 1]$.
We propose an approach to address the challenges of DDA and hard negative samples, as shown in Figure \ref{fig:architecture}.
The approach improves the ability of the model to measure the semantic similarity of the text.
We modify the self-attention layers of the Transformer encoder (BERT) and correct the attention scores by external matrices.
We call this approach IEKM.

\subsection{The Matrices}

Take the example of sentence pairs $s_{1}$ and $s_{2}$, we construct a similarity matrix $M$ and a dissimilarity matrix $M_r$ based on the two input sentences, both of size $(l_{1} + l_{2}) \times (l_{1} + l_{2})$, where $l_{1}$ and $l_{2}$ are the lengths of the two sentences, respectively.

% For the similarity matrix $M$, the score of each cell is calculated by the word similarity tools Cilin and Hownet.
In the training phase, to provide supervised information, for the similarity matrix $M$, the score of each cell between sentence pairs is calculated by the word similarity tools (Cilin and Hownet).
% Specifically, in the computation first use a segmentation tool, such as Jieba\footnote{\url{https://github.com/fxsjy/jieba}}, to segment the sentence pairs.
Specifically, a segmentation tool, such as Jieba\footnote{\url{https://github.com/fxsjy/jieba}}, is first used to segment the sentence into words before the calculation.
Then, the words similarity scores between sentence pairs are calculated word by word using the word similarity tools mentioned above.
Since our goal is to correct the self-attention scores between sentence pairs $s_{1}$ and $s_{2}$ and not the self-attention scores between a sentence itself and itself, we only calculate the words similarity score of $\omega_{a}$ and $\omega_{b}$, where $\omega_{a}$ and $\omega_{b}$ are the words of $s_{1}$ and $s_{2}$, respectively.
The similarity score $c_{ab}$ of $\omega_{a}$ and $\omega_{b}$ in $M$ is calculated as follows.

% \begin{small}
\begin{equation}
c_{ab}=\left\{\begin{array}{l}
0 \qquad\qquad\qquad\qquad\quad\;\;\text{if } w_{a}, w_{b}\in {s^\prime}, \\
\frac{\operatorname{Lin}(w_{a}, w_{b})+How(w_{a}, w_{b})}{2} \quad\text{othewise},
\end{array}\right.
\label{formula:cab}
\end{equation}
% \end{small}

where $Lin(\cdot)$ and $How(\cdot)$ denote the tools Cilin and Hownet, respectively.
$w_{a}, w_{b}\in {s^\prime}$ denote $w_{a}$ and $w_{b}$ come form the same sentence.
Here, a larger score of $c_{ab}$ indicates that the word $\omega_{a}$ is more similar to $\omega_{b}$, and vice versa.
To accommodate models with token-based input, such as BERT, after obtaining the similarity scores between words, we make copies in token units to obtain the similarity scores between tokens, i.e., the score of each cell of the matrix $M$.
An example of a similarity matrix $M$ is shown in Figure \ref{fig:ex_matrix}.
% 需要注意的是，为了适应以字为单位的输入，我们将词之间的分数以字数拷贝

The $c_{ab}^{r}$ of dissimilarity matrix $M_{r}$ can be obtained directly by matrix manipulation using the similarity matrix $M$ in this paper, as follows:
% \begin{equation}
% M_{r} = 1 - M.
% \end{equation}
% \begin{small}
\begin{equation}
c_{ab}^{r}=\left\{\begin{array}{l}
0 \qquad\quad\;\,\text {if } w_{a}, w_{b}\in{s^\prime}, \\
1-c_{ab} \quad\text{othewise}.
\end{array}\right.
\label{formula:cab}
\end{equation}
% \end{small}

\subsection{IEKM}

IEKM is based on BERT (the encoder of the Transformer) \cite{DBLP:conf/naacl/DevlinCLT19} for modification.
The main difference between BERT and IEKM is the self-attention layer, as shown in Figure \ref{fig:Our self-attention}.

Their frameworks are a combination of multi-layer Multi-head attention and Feed Forward.
BERT’s attention function can be described as a mapping from query vector $Q$ and a set of key-value vector pairs ($K$, $V$) to an output vector–the attention strengths. 
Multi-head attention linearly projects the queries, keys and values $h$ times ($h$ is the number of “heads”) with different linear projections to $d_{k}$, $d_{k}$,and $d_{v}$ dimensions,respectively \cite{DBLP:conf/nips/VaswaniSPUJGKP17}.

IEKM needs to incorporate the similarity matrix $M$ and dissimilarity matrix $M_{r}$ in calculating the self-attention scores for correction.
The matrix incorporation operation is to first perform a Hadamard product of the matrices $M$ and $M_{r}$ with the scores of self-attention, respectively, and then add them all three.
The learnable gate control unit $g$ is used to control the exact amount of information obtained from the matrices $M$ and $M_{r}$, respectively.
In this way when faced with an actual sentence pair, the model can automatically adjust whether to pay more attention to similar words or irrelevant words.
The calculation formula is shown below:
% \begin{small}
\begin{equation}
% \begin{gathered}
\begin{aligned}
& \text{MultiHead}(Q, K, V, G)= \\
& \qquad \text{Concat}\left(\text{head}_1,\ldots,\text{head}_h\right) W^O, \\
& \text {head}_i= \\
& \qquad \text{Attention}\left(QW_i^Q, KW_i^K, VW_i^V, GW_i^G\right), \\
\end{aligned}
\end{equation}
where $W_i^Q\in\mathbb{R}^{d_{m} \times d_{k}}$, $W_i^K\in\mathbb{R}^{d_{m} \times d_{k}}$, $W_i^V\in\mathbb{R}^{d_{m} \times d_{v}}$, $W_i^O\in\mathbb{R}^{h d_{v} \times d_{m}}$ and $W_i^G\in\mathbb{R}^{d_{m} \times d_{k}}$ are parameter matrices representing the projections. 
$d_{m}$ is model’s hidden states size. 
The calculation of IEKM’s self-attention is as follows:
% \begin{small}
\begin{equation}
% \begin{gathered}
\begin{aligned}
& \operatorname{Attention}(Q, K, V, G)= \\
& \qquad \operatorname{softmax}\left(\frac{p}{\sqrt{d_{k}}} + MASK \right) V, \\
& p = {Q K^{T}} \odot \lbrack 1 + g \odot M + (1-g) \odot M_{r} \rbrack , \\
% & \qquad {Q K^{T}} \odot \lbrack 1 + g \odot M + (1-g) \odot M_{r} \rbrack , \\
& g = Linear(G) = G W_{l}^{T} + b,
% \end{gathered}
\end{aligned}
\end{equation}
% \end{small}

where $MASK$ is a matrix used in masked language modeling.
The matrix $G$ is similar to the matrices Q,K,V obtained from the input data over the Linear layer.
$W_{l} \in\mathbb{R}^{d_{k} \times 1}$ is the parameter matrix representing the projections.
% $d_{k}$ denotes the dimension that the input matrices project to each head \citep{DBLP:conf/nips/VaswaniSPUJGKP17}.
Therefore, each head has its learnable gate control unit, and they can learn different dimensions of information to control the input of similarity and dissimilarity matrices.

\section{Experiments}
In this section we evaluate our approach from two aspects.
First, we use four STS datasets to train and test the IEKM model separately, without using the domain dictionary.
% Among them, the LOGS test set is out-of-domain of the TCXSS dataset's train set.
% The LOGS test set of the TCXSS consists of hard negative samples and multi-domain log data.
Secondly, to demonstrate that our approach can solve the hard negative samples and DDA challenges without retraining the model, we perform a flexible correction study, i.e., construct a domain dictionary for the LOGS test set and conduct experiments
% The results show that by constructing an external dictionary, the model performance is greatly improved and can be adapted to different domains.

\begin{table}[!htb]
% \begin{table}[!]
\begin{center}
  \begin{tabular}{lccc}
    \toprule
    Dataset & Train & \multicolumn{2}{c}{Test} \\
    \midrule
    TUC & 42,220 & \multicolumn{2}{c}{9334} \\
    LCQMC & 238,766 & \multicolumn{2}{c}{12,500} \\
    BQ Corpus & 100,000 & \multicolumn{2}{c}{10,000} \\
    \midrule
    & Train & Test-FIT & Test-LOGS \\
    \midrule
    TCXSS & 295,067 & 101,187 & 2,373 \\    %3k len:3376 education, first_eva len:19613 customer service, fit len:101187 financial
    \bottomrule
  \end{tabular}
  \caption{The statistics of datasets TUC, LCQMC, BQ Corpus and TCXSS.}
  \label{table:datasets}
% \end{table}
\end{center}
\end{table}

% \begin{table}[htb]
\begin{table*}[!]
\begin{center}
% \begin{small}
  \begin{tabular}{l|c|c|c|cc}
    \toprule
     & \multicolumn{1}{c|}{TUC} & \multicolumn{1}{c|}{LCQMC} & \multicolumn{1}{c|}{BQ Corpus} & \multicolumn{2}{c}{TCXSS} \\
     & (F1) & (ACC) & (ACC) & FIT (ACC) & LOGS (F1) \\
    \midrule
    BERT-base & $76.0^{\spadesuit} $ & 86.1$^{\clubsuit}$ & 84.6$^{\clubsuit}$ & 79.74 & 50.54 \\
    BERT-Sim & \textbf{76.2}$^{\spadesuit}$ & 86.50 & 84.19 & 79.82 & 50.13 \\
    IEKM-base & 76.19 & \textbf{87.09} & \textbf{84.71} & \textbf{80.00} & \textbf{56.61} \\
    \midrule
    BERT-large & 76.15 & 86.8$^{\clubsuit}$ & 84.9$^{\clubsuit}$ & 80.32  & 51.85 \\
    IEKM-large & \textbf{76.45} & \textbf{87.12} & \textbf{85.48} & \textbf{80.84} & \textbf{57.37} \\
    \bottomrule
  \end{tabular}
  \caption{The SSEI performance on in three datasets.
We highlight the highest numbers among the results.
$\clubsuit$: results from \url{https://github.com/ymcui/Chinese-BERT-wwm}. $\spadesuit$: results from BERT-Sim \citep{DBLP:conf/www/XiaWTC21}.
  }
  \label{table:main_results}
% \end{small}
\end{center}
\end{table*}

\begin{table}[!h]
\begin{center}
  \begin{tabular}{lc}
    \toprule
    & LOGS (F1) \\
    \midrule
    BERT-base & 50.54 \\
     + IEKM & 56.61 \\
     + IEKM + keywords & \textbf{73.53} \\
    \midrule
    Number of keyword pairs added & 33 \\
    % Number of samples with keyword pair & 113 \\
    Number of affected samples & 113 \\
    \bottomrule
  \end{tabular}
  \caption{The results of flexibility correction study. There are 33 keyword pairs in the dictionary, and 113 sentence pairs in the test set contain a keyword pair from the dictionary.}
  \label{table:flexibility correction}
\end{center}
\end{table}

\subsection{Datasets}
We conduct experiments on two types of datasets separately: the training set and the test set belong to the same domain or different domains. 
A brief description of these datasets is summarized in Table \ref{table:datasets}. 
Below we briefly describe each dataset.

\textbf{Twitter Url Corpus (TUC)} \cite{DBLP:conf/emnlp/LanQHX17} is an English STS dataset collected from tweets that share the same URL of news articles. 
It includes 56,787 sentence pairs, and each sentence pair is annotated by 6 Amazon Mechanical Turk workers. 
If $n \leq 2$ workers are positive, we treat them as non-paraphrasing; if $n \geq 4$, we treat them as paraphrasing; if $n = 3$, we discard them.
% After this treatment, there are 42,220 pairs for training and 9,334 pairs for test.

\textbf{LCQMC} \citep{DBLP:conf/coling/LiuCDZCLT18} and \textbf{BQ Corpus} \citep{DBLP:conf/emnlp/ChenCLYLT18} are the large-scale Chinese question-matching corpus. 
% LCQMC is more general than the paraphrase corpus as it focuses on intent matching rather than paraphrasing.
They are binary classification task datasets that allow the model to learn to determine whether a sentence pair represents the same meaning

% \textbf{BQ Corpus} \citep{DBLP:conf/emnlp/ChenCLYLT18} is the Bank Question (BQ) corpus, a large-scale domain-specific Chinese corpus for SSEI (sentence semantic equivalence identification).
% Also, BQ is a binary classification task dataset.

% \textbf{TCXSS (Tencent Cloud Xiaowei sentence similarity dataset)} is a multi-domain dataset.
\textbf{TCXSS (a sentence similarity dataset\footnote{The full dataset name may indirectly reveal affiliation so only abbreviations are used here.})} is a multi-domain dataset.
Unlike the above three datasets, TCXSS is a regression task dataset, which means that a sentence pair has a label that takes a score between 0 and 1.
This label measures the similarity of the sentence pair (the smaller the score, the more similar the sentence pair). 
The training data is mainly from the generic domain. 
But the testing data consists of actual industrial data from different domains: the financial domain (FIT), and the user logs (LOGS).
LOGS mainly contains negative negative samples and multi-domain data, e.g. [\begin{CJK*}{UTF8}{gbsn}\begin{small}初级考试成绩在哪里查询\end{small}, \begin{small}注会考试成绩在哪里查询\end{small}, Negative\end{CJK*}] ([Where to check the junior exam results, Where to check the CPA exam results, Negative]), [\begin{CJK*}{UTF8}{gbsn}\begin{small}怎么得到社保卡\end{small}, \begin{small}怎么拥有社保卡\end{small}, Positive\end{CJK*}] ([How to get a social security card, How to have a social security card, Positive]).
Therefore, the LOGS test set can better reflect the performance of the model in the face of hard negative samples and DDA challenges.
% Each sample of the test data consists of one query and 20 candidates, from which the model is required to find the candidate with the closest meaning to the query.

\subsection{Evaluation Metrics}
Accuracy (ACC) the widely-used STS metric is utilized to evaluate our method.
But, for TUC and LOGS test set in the TCXSS, we use F1 score to evaluate the performance of our method because of the uneven label distribution.
A higher value of these metrics indicates better STS performance.

% \subsection{Experimental details}
% We implement the models with the same PyTorch framework.
% In this paper, we use chinese-roberta-wwm-ext \citep{cui2019pre} and bert-base-uncased \cite{DBLP:conf/naacl/DevlinCLT19} as our pre-trained model to initialize the parameters of BERT and IEKM respectively, and then fine-tune them.
% To get better performance, we use cross-encode method to encode the data \cite{DBLP:conf/naacl/QuDLLRZDWW21}.
% The input data format is similar to BERT, we split the input sentence pair with a special [SEP] token.
% We also insert a special [CLS] token in the front of the input sequence and append a [SEP] token at the end of the input sequence.
% We input the hidden states vector of the [CLS] output to the Dsense layer and then to the classification layer to get the final output.
% We train the model for 5 epochs and we set the learning rate as 2e-5.
% And we set the batch size to 64 for all datasets (when using large model, we set the batch size to 32).
% For the dataset TCXSS, in the testing phase we set the IEKM' threshold to 0.326, i.e., less than this threshold means that the sentence pair is positive and the opposite is negative.
% % Our PyTorch implementation is available at: https://github.com/wulaoshi/IEKM.
% Our PyTorch implementation will be available at: https://github.com/***.

\subsection{Main Results}
In theory, all models based on the Transformer backbone can use our method, so we choose BERT for our experiments.
Table \ref{table:main_results} shows the results of our implementation.
We conduct experiments on English dataset and Chinese dataset respectively. 
We also compare with BERT and BERT-Sim \cite{DBLP:conf/www/XiaWTC21} (an advanced model incorporating external similar word information).
From the LOGS results, it can be inferred that for hard negative samples and DDA data, our method still has very good performance even if the external matrices are constructed with the word similarity tools only.
This is consistent with our goal: by constructing the IEKM attention matrix, the model results are corrected to the ones we want.
And the subsequent experiment will also prove this conclusion.

\subsection{Flexibility correction study} \label{Flexibility correction study}

The core value of our approach is to flexibly control the output of the model by inputting external keyword information to cost-effectively solve the challenges of hard negative samples and DDA.
The IEKM attention mechanism has a shot at incorporating external matrices to get more information to correct the model results.
At the same time, the gate control unit learns how much external information needs to be incorporated based on the input data during the training phase.

A keyword dictionary is built for the LOGS test set to simulate the real usage scenarios of our customers.
During the test, input data [$s_{1}$, $s_{2}$] is replaced with [$s_{1}$, $s_{2}$, $keyword_{1}$, $keyword_{2}$, $score$], where $keyword_{1}$, $keyword_{2}$ are the words in sentences $s_{1}$ and $s_{2}$ respectively, and the $score \in \{0, 1\}$ is the correlation coefficient of these two words in the keyword dictionary, which is originally calculated by Equation \ref{formula:cab}.
For example, change the input data [\begin{CJK*}{UTF8}{gbsn}\begin{small}收益哪里看\end{small}, \begin{small}在哪里查看总收益\end{small}\end{CJK*}] ([Where to see earnings, Where to see total earnings]) to [\begin{CJK*}{UTF8}{gbsn}\begin{small}收益哪里看\end{small}, \begin{small}在哪里查看总收益, 收益，总收益\end{small}，0\end{CJK*}] ([Where to see earnings, Where to see total earnings, earnings, total earnings, 0]).
The results are shown in Table \ref{table:flexibility correction}, and Table \ref{table:examples} shows the cases study.

It is a flexible way to correct the model results by adding similar word matrices.
Rather than directly influencing the model's results, this approach informs the model which tokens we care more about and then allows the model to make reasonable judgments based on the context.
The results in Table \ref{table:flexibility correction} show that when facing hard negative samples and DDA challenges, we only need to build a dictionary of relevant keywords for IEKM to solve the challenges (no need to label data to train new models and no need to deploy multiple models).
Also, the results in Table \ref{table:examples} show that our method is a flexibility correction approach.

\subsection{Ablation Studies}
We investigate the impact of incorporating different matrices.
All reported results in this section are based on the LOGS test set.

\textbf{+ M.} Only the similarity matrix is used.
The Hadamard product operation is performed using the matrix of self-attention with the similarity matrix.

\textbf{+ Mr.} Only the dissimilarity similarity matrix is used.
The Hadamard product operation is performed using the matrix of self-attention with the similarity matrix.  

\textbf{+ M + Mr + gate.} Both similarity and dissimilarity matrixes are used and their weights are adjusted by using a learnable gate control unit.

Table \ref{table:ablation studies} shows the results of the ablation studies.

\begin{table}[!ht]
\begin{center}
  \begin{tabular}{lccl}
    \toprule
    & logs (F1) \\
    \midrule
    BERT-base & 50.54 \\
     + M & 51.3 \\
     + Mr & 52.83 \\
     + M + Mr + gate & \textbf{56.61} \\
    \bottomrule
  \end{tabular}
  \caption{The results of ablation studies.}
  \label{table:ablation studies}
\end{center}
\end{table}

\section{Conclusion}
In this work, to address two common challenges of customer service platform systems in industrial scenario, i.e., negative negative samples and DDA, we propose IEKM, a method that can incorporate external similarity word matrices, which can flexibly change the model results by receiving external similarity words information.
% We further demonstrate the effectiveness of the method through experiments and analysis.
We further demonstrate the effectiveness of the method through experiments.
By feeding an external dictionary, the method enables the model to address the above challenges without the need to retrain the model by re-labeling the data, and without the need to deploy multiple models.
We believe that it can have a broader application in industry, especially for scenarios where there is a requirement for multi-domain.

\bibliography{anthology,custom}
\bibliographystyle{acl_natbib}

% \newpage
\clearpage
\appendix
\label{sec:appendix}

\section{Experimental details}
We implement the models with the same PyTorch framework.
In this paper, we use chinese-roberta-wwm-ext \citep{cui2019pre} and bert-base-uncased \cite{DBLP:conf/naacl/DevlinCLT19} as our pre-trained model to initialize the parameters of BERT and IEKM, and then fine-tune them.
To get better performance, we use cross-encode method to encode the data \cite{DBLP:conf/naacl/QuDLLRZDWW21}.
The input data format is similar to BERT, we split the input sentence pair with a special [SEP] token.
We also insert a special [CLS] token in the front of the input sequence and append a [SEP] token at the end of the input sequence.
We input the hidden states vector of the [CLS] output to the Dsense layer and then to the classification layer to get the final output.
We train the model for 5 epochs and set the learning rate as 2e-5.
And we set the batch size to 64 for all datasets (when using large model, we set the batch size to 32).
For the dataset TCXSS, in the testing phase we set the IEKM' threshold to 0.326, i.e., less than this threshold means that the sentence pair is positive and the opposite is negative.
% Our PyTorch implementation is available at: https://github.com/wulaoshi/IEKM.
Our PyTorch implementation will be available at: https://github.com/***.

\section{Analysis}
In this section, we conduct further analyses to explain the reasons why IEKM can achieve flexible correction results.

\textbf{Similar Words Supervised Signal.} 
In the real world, the similarity between keywords in some domains usually greatly affects the similarity of the sentence pair \citep{DBLP:journals/corr/abs-2003-11516}. 
Whether it is WordNet \citep{DBLP:journals/cacm/Miller95} in English or Hownet in Chinese, they can calculate the similarity coefficients between words.
% These similarity coefficients serve as a supervised signal to match each other with sentence pairs.
In the training phase, these similarity coefficients serve as supervised signals that enable the model to learn the relationship between similar word information and sentence similarity.
Thereby, in the inference stage, the final sentence similarity result can be adjusted according to the external similarity word coefficients.

\textbf{Integration of external information matrices through gate control units.} There are many ways to incorporate external information into the model, but the IEKM method is a perfect fit for this purpose. 
Firstly, we construct the external information as a bunch of words coefficients and joint them into a coefficient matrix, which is consistent with the coefficient matrix computed by the self-attention layer in Transformer, and thus they can be directly manipulated for matrix computation. 
Secondly, learnable gate control units are used to control the different layers of the model to automatically draw external information according to the context. 
This approach flexibly corrects the model results and makes the model more robust when faced with more challenges. 
Finally, in the inference phase, we modify the external matrix based on industry dictionaries to inform the model which words are similar and which words are irrelevant, thus making the model results corrected in our desired direction.

\section{Cases Study}
\setcounter{figure}{0}
\setcounter{table}{0}
\renewcommand{\thetable}{C.\arabic{table}}
\renewcommand{\thefigure}{C.\arabic{figure}}
\begin{figure*}[htb]
  \centering
  \includegraphics[width=0.8\textwidth]{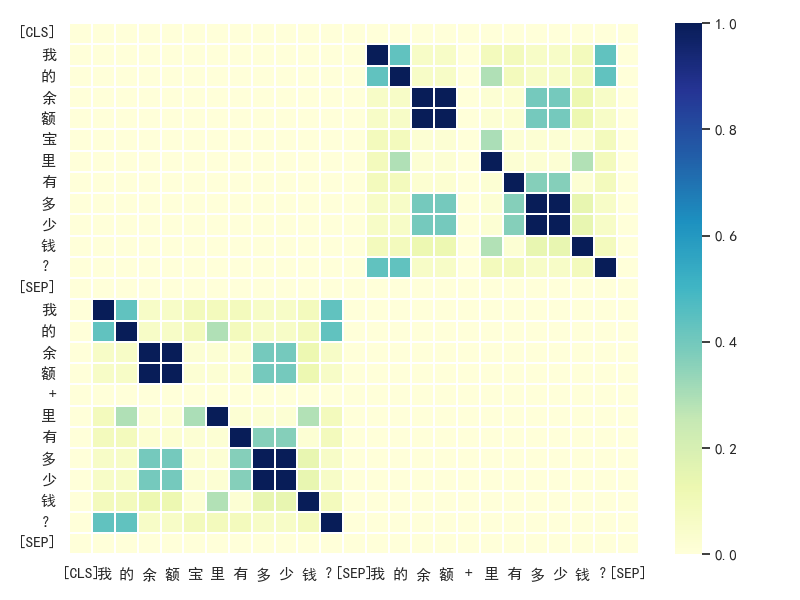}
  \caption{An example of a similarity matrix $M$. The darker the color of the cell, the larger the score, and the more similar the corresponding tokens are to each other.}
  \label{fig:ex_matrix}
\end{figure*}
We visualize a matrix M of an input sentence pair, as shown in Figure \ref{fig:ex_matrix}.
For the section \ref{Flexibility correction study}, we have added a case study and the results are shown in the Table \ref{table:examples}.
The results show that by adding different dictionaries, IEKM's output results also change. 
The change trend is in line with our target, but the change value of the result is calculated by IEKM based on the context. 
This is the performance of the flexibility correction.

\begin{CJK*}{UTF8}{gbsn}
\begin{table*}[htb]
\begin{center}
    \begin{tabular}{ll|c|c} 
    \hline
    \multicolumn{2}{c|}{\multirow{2}{*}{Sentence piar}} & \multicolumn{1}{c|}{\multirow{2}{*}{Add keywords}} & (BERT-base + \\
    & & & IEKM) scores  \\ 
    \midrule
    \multirow{2}{*}{\begin{small}{收益哪里看}\end{small}} & \multirow{2}{*}{\begin{small}{在哪里查看总收益}\end{small}} & $\varnothing$ & 0.241 \\
     &  & {\begin{small}收益，总收益\end{small}，0} & 0.572 \\
    \begin{small}{(Where to see earnings)}\end{small} & \begin{small}{(Where to see total}\end{small} & {\begin{small}收益，总收益\end{small}，1} & 0.195 \\
     & \begin{small}{earnings)}\end{small} & (earnings, total earnings) & \\
    % \begin{small}{}\end{small} &  &  & \\    
    \midrule
    \multirow{2}{*}{\begin{small}{刚买的网课想撤销}\end{small}} & \multirow{2}{*}{\begin{small}{我想取消买的教材}\end{small}} & $\varnothing$ & 0.461 \\
     &  & {\begin{small}网课，教材，0\end{small}} & 0.529 \\
     \begin{small}{(Cancel the online}\end{small} & \begin{small}{(I want to cancel the}\end{small} & {\begin{small}网课，教材\end{small}，1} & 0.261 \\
     \begin{small}{lessons I just bought)}\end{small} & \begin{small}{textbook I bought)}\end{small} & (online lessons, textbook) & \\
     % &  &  & \\

    \midrule
    \multirow{2}{*}{\begin{small}{社保卡领取}\end{small}} & \multirow{2}{*}{\begin{small}{社保卡申领}\end{small}} & $\varnothing$ & 0.161 \\
     &  & {\begin{small}领取，申领\end{small}，0} & 0.448 \\
    \begin{small}{(Receiving social}\end{small} & \begin{small}{(Applying for social }\end{small} & {\begin{small}领取，申领\end{small}，1} & 0.033 \\
    \begin{small}{security card)}\end{small} & \begin{small}{security card)}\end{small} & (Receiving, Applying) & \\
     % &  &  & \\
    
    \midrule
    \multirow{2}{*}{\begin{small}{我能不能升职}\end{small}} & \multirow{2}{*}{\begin{small}{我能不能升舱}\end{small}} & $\varnothing$ & 0.587 \\
     &  & {\begin{small}升职，升舱\end{small}，0} & 0.741 \\
    \begin{small}{(Can I get a promotion)}\end{small} & \begin{small}{(Can I upgrade)}\end{small} & {\begin{small}升职，升舱\end{small}，1} & 0.193 \\
     &  & (promotion, upgrade) & \\
    
    \midrule
    \multicolumn{3}{c|}{Percentage of hard-negative samples resolved after adding the keyword dictionary} & 85.5\% \\
    \bottomrule
  \end{tabular}
  \caption{Flexibility correction cases. The results show that the flexible correction makes the model results change in the direction we expect, but the value of the change is calculated by the model itself.
  }
  \label{table:examples}
\end{center}
\end{table*}
\end{CJK*}

\end{document}